# PolyNet: A Pursuit of Structural Diversity in Very Deep Networks

Xingcheng Zhang[*]    Zhizhong Li[*]    Chen Change Loy    Dahua Lin

{zx016, lz015, ccloy, dhlin}@ie.cuhk.edu.hk

## Abstract

*A number of studies have shown that increasing the depth or width of convolutional networks is a rewarding approach to improve the performance of image recognition. In our study, however, we observed difficulties along both directions. On one hand, the pursuit for very deep networks is met with a diminishing return and increased training difficulty; on the other hand, widening a network would result in a quadratic growth in both computational cost and memory demand. These difficulties motivate us to explore structural diversity in designing deep networks, a new dimension beyond just depth and width. Specifically, we present a new family of modules, namely the PolyInception, which can be flexibly inserted in isolation or in a composition as replacements of different parts of a network. Choosing PolyInception modules with the guidance of architectural efficiency can improve the expressive power while preserving comparable computational cost. The Very Deep PolyNet[1], designed following this direction, demonstrates substantial improvements over the state-of-the-art on the ILSVRC 2012 benchmark. Compared to Inception-ResNet-v2, it reduces the top-5 validation error on single crops from 4.9% to 4.25%, and that on multi-crops from 3.7% to 3.45%.* [2]

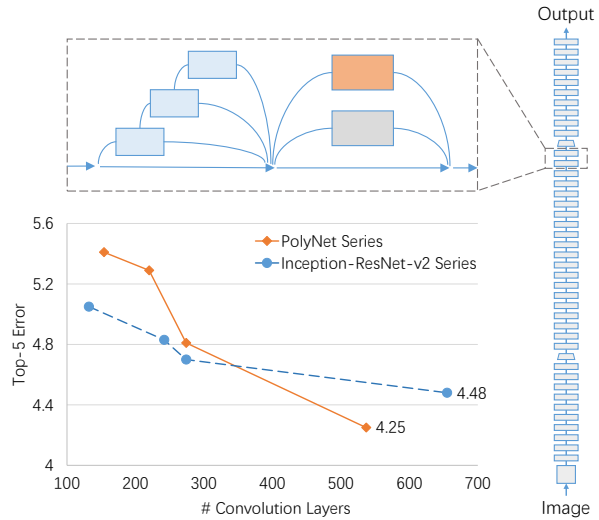

Figure 1: We propose a new family of building blocks called PolyInceptions. PolyNet, a new network architecture using PolyInceptions, outperforms Inception-ResNet-v2 and a deepened version thereof, reducing the single-crop top-5 error to 4.25% on ILSVRC 2012 validation set.

## 1. Introduction

The past four years witnessed a dramatic reduction of classification error on the ILSVRC benchmark [19]. Along with this progress is a remarkable increase in network depth – from AlexNet [16] to ResNet [9], the numbers of layers have been raised from 8 to 152. This strong correlation between performance and depth has motivated a series of work [10, 21] to pursue very deep networks. On the other hand, increasing the width of a network has also been shown to be an effective way to improve performance [1, 28, 33]. This line of exploration also draws lots of attention recently.

*Is increasing the depth or width the best strategy to pursue optimal performance?* Whereas a positive answer to this question seems to be backed by numerous evidences in previous work, our empirical study suggested otherwise. Specifically, we found that as we continue to increase the depth, the reward is diminishing beyond a certain level. Another direction, namely, widening a network, also has its own problems. The most significant problem is the quadratic growth in both runtime complexity and memory demand. Our study showed that this significant increase in computational cost may not be justified by the performance gain that it brings.

These issues motivate us to explore *structural diversity*, an alternative dimension in network design. Actually, the significance of diversity has been repeatedly demonstrated in previous work, such as Inception modules [26], network ensembles [15], and a recent study [30] that relates ResNet with an implicit ensemble. However, several questions remain yet to be answered: (1) what is the best way to build diversity into a network, and (2) when the pursuit of diver-sity is a more rewarding strategy than the pursuit of depth.

---
[*]These two authors contribute equally to the work.
[1] Codes available at https://github.com/CUHK-MMLAB/polynet.
[2]This paper is accepted by CVPR 2017.

In this work, we propose a new family of modules called *PolyInception*. Specifically, each PolyInception module is a *"polynomial"* combination of Inception units, which integrates multiple paths in a parallel or cascaded fashion, as illustrated in Fig. 1. Compared to the conventional paradigm where basic units are stacked sequentially, this alternative design can substantially enhance the structural diversity of a network. Note that the PolyInception modules are designed to serve as building blocks of large networks, which can be inserted in isolation or in a composition to replace existing modules. We also present an ablation study that compares the performances and costs of various module configurations under different conditions, which provides useful guidance for network design.

Based on a systematic comparison of different designing choices, we devise a new network called *Very Deep PolyNet*, which comprises three stages operating on different spatial resolutions. PolyInception modules of different configs are carefully chosen for each stage. On the ILSVRC 2012 validation set, the Very Deep PolyNet achieves a top-5 error rate $4.25\%$ on single-crop and $3.45\%$ on multi-crop. Compared to Inception-ResNet-v2 [25], the latest state-of-the-art which obtains top-5 error rates $4.9\%$ (single-crop) and $3.7\%$ (multi-crop), this is a substantial improvement. Furthermore, Very Deep PolyNet also outperforms a deepened version of Inception-ResNet-v2 under the same computational budget, with a considerably smaller parameter size ($92M$ vs. $133M$).

Overall, the major contributions of this work consist in several aspects: (1) a systematic study of *structural diversity*, an alternative dimension beyond depth and width, (2) *PolyInception*, a new family of building blocks for constructing very deep networks, and (3) *Very Deep PolyNet*, a new network architecture that obtains best single network performance on ILSVRC.

## 2. Related Work

The *depth revolution* is best depicted by the various successful convolutional neural networks in the ImageNet challenge [19]. These networks, from AlexNet [16], VGG [21], GoogleNet [26], to ResNet [9], adopted increasingly deeper architectures, However, very deep networks are difficult to train, prone to overfitting, and sometimes even leading to performance degradation, as reported in [2, 6, 8, 23, 24]. Residual structures [9, 10] substantially mitigate such issues. One of the reason is that it introduces identity paths that can efficiently pass signals both forward and backward. Techniques like BN [14] and dropout [23] also facilitate the training process. In addition, extensive studies [7, 8, 18, 27] have been done to find good design choices, *e.g.* those in depth, channel sizes, filter sizes, and activation functions.

Some recent studies challenged that depth is not the secret ingredient behind the success of ResNet. One of the arguments is that more powerful residual blocks with enhanced representation capabilities can also achieve similar or even better performance without going too deep. Wide residual networks (WRN) [33] increases width while reducing depth. Zhang *et al.* [34] propose ResNet of ResNet (RoR). Targ *et al.* [28] propose ResNet in ResNet (RiR), which generalizes residuals to allow more effective learning. Multi-residual networks [1] extends ResNet in a different way by combining multiple parallel blocks.

From another perspective, the residual structure can be seen as an exponential ensemble of relatively shallow networks. Veit *et al.* [30] hypothesized that the power lies in this ensemble. Networks that utilize this implicit ensemble are called *ensemble by structure*. One can also encourage the formation of ensemble during the learning process, which is called *ensemble by train*. Stochastic depth technique [13] randomly drops a subset of layers in training. The process can be interpreted as training an ensemble of networks with different depths. This way not only eases the training of very deep networks, but also improves performance. Swapout [22] is a stochastic training method that generalizes dropout and stochastic depth by using random variables to control whether units are kept or dropped. Some recently designed networks [1, 17, 22] embrace this ensemble interpretation.

Shortcut paths in residual networks and Highway networks [24] inspire explorations on more general topological structures rather than the widely adopted linear formation. Two of such examples are DenseNet [12], which adds shortcuts densely to the building blocks, and FractalNet [17], which uses a truncated fractal layout.

Inception-ResNet-v2 [25] achieved so far the best published single model accuracy on ImageNet to our knowledge. It is a combination of the latest version of Inception structure and the residual structure, in which Inception blocks are used to capture the residuals. Inception blocks are the fundamental components of GoogleNet [26]. Their structures have been optimized and fine-tuned for several iterations [25, 26, 27]. We adopt Inception-ResNet-v2 as the base model in our study.

## 3. PolyInception and PolyNet

*Depth*, the number of layers, and *width*, the dimension of intermediate outputs, are two important dimensions in network design. When one wants to improve a network's performance, going deeper and going wider are the most natural choices provided that the computational budget allows. The ultimate goal of this study is to push forward the state of the art of deep convolutional networks. Hence, like many others, we begin with these two choices.

As mentioned, the benefit of going deeper has been clearly manifested in the evolution of deep network archi-tectures over the past several years [9, 10, 21, 26]. Following this path, we continue to stack residual units on top of ResNet-152. As shown in Fig. 2, we can still observe significant performance gain when the number of residual units is increased from 16 (ResNet-50) to 89 (ResNet-269). Yet, as we further increase the number to 166 (ResNet-500), the reward obviously diminishes. The cost doubles, while the error rate decreases from 4.9% to 4.8%, only by 0.1%. Clearly, the strategy of increasing the depth becomes less effective when the depth goes beyond a certain level. The other direction, namely going wider [33], also has its own issues. When we increase the numbers of channels of all layers by a factor of $k$, both the computational complexity and the memory demand increases by a factor of $k^2$. This quadratic growth in computational cost severely limits the extent one can go along this direction.

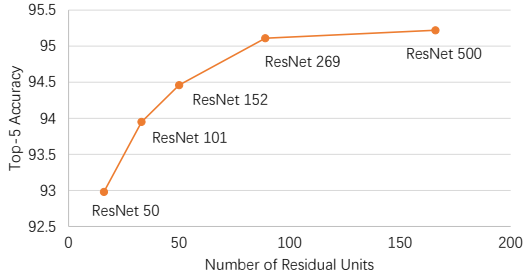

Figure 2: Top-5 single crop accuracies of ResNets [10] with different number of residual units on the ILSVRC 2012 validation set. As the number of residual units increases beyond 100, we can see that the return diminishes.

The difficulties met by both directions motivate us to explore alternative ways. Revisiting previous efforts in deep learning, we found that *diversity*, another aspect in network design that is relatively less explored, also plays a significant role. This is manifested in multiple ways. First, *ensembles* usually perform better than any individual networks therein – this has been repeatedly shown in previous studies [3, 5, 11, 20]. Second, as revealed by recent work [13, 30], ResNets, which have obtained top performances in a number of tasks, can be viewed as an implicit ensemble of shallow networks. It is argued that it is the *multiplicity* instead of the *depth* that actually contributes to their high performance. Third, Inception-ResNet-v2, a state-of-the-art design developed by Szegedy *et al*. [25], achieves considerably better performance than ResNet with substantially fewer layers. Here, the key change as compared to ResNet is the replacement of the simple residual unit by the Inception module. This change substantially increases the structural diversity within each unit. Together, all these observations imply that *diversity*, as a key factor in deep network design, is worthy of systematic investigation.

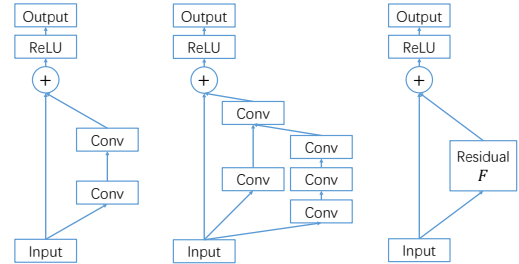

Figure 3: **Left**: residual unit of ResNet [9]. **Middle**: type-B Inception residual unit of Inception-ResNet-v2 [25]. **Right**: abstract residual unit structure where the residual block is denoted by $F$.

### 3.1. PolyInception Modules

Motivated by these observations, we develop a new family of building blocks called *PolyInception modules*, which generalize the Inception residual units [25] via various forms of *polynomial compositions*. This new design not only encourages the structural diversity but also enhances the expressive power of the residual components.

To begin with, we first revisit the basic design of a *residual unit*. Each unit comprises two paths from input to output, namely, an *identity path* and a *residual block*. The former preserves the input while the latter turns the input into residues via nonlinear transforms. The results of these two paths are added together at the output end of the unit, as

$$(I + F) \cdot \mathbf{x} = \mathbf{x} + F \cdot \mathbf{x} := \mathbf{x} + F(\mathbf{x}). \quad (1)$$

This formula represents the computation with operator notations: $\mathbf{x}$ denotes the input, $I$ the identity operator, and $F$ the nonlinear transform carried out by the residual block, which can also be considered as an operator. Moreover, $F \cdot \mathbf{x}$ indicates result of the operator $F$ acting on $\mathbf{x}$. When $F$ is implemented by an Inception block, the entire unit becomes an *Inception residual unit* as shown in Fig. 3. In standard formulations, including ResNet [9] and Inception-ResNet [25], there is a common issue – the residual block is very shallow, containing only 2 to 4 convolutional layers. This restricts the capacity of each unit, and as a result, it needs to take a large number of units to build up the overall expressive power. Whereas several variants have been proposed to enhance the residual units [1, 28, 34], the improvement, however, remains limited.

Driven by the pursuit of structural diversity, we study an alternative approach, that is, to generalize the additive combination of operators in Eq. (1) to *polynomial compositions*. For example, a natural extension of Eq. (1) along this line is to simply add a second-order term, which would result in a new computation unit as

$$(I + F + F^2) \cdot \mathbf{x} := \mathbf{x} + F(\mathbf{x}) + F(F(\mathbf{x})). \quad (2)$$

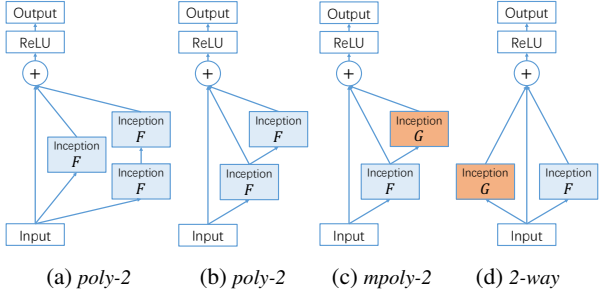
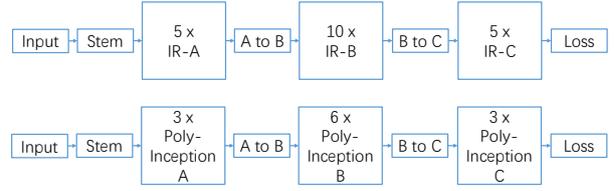

(a) *poly-2*   (b) *poly-2*   (c) *mpoly-2*   (d) *2-way*

Figure 4: Examples of PolyInception structures.

Figure 5: **Upper**: The overall schema of Inception-ResNet-v2. Adapted from [25]. *IR-A*, *IR-B* and *IR-C* are three types of Inception residual units designed for stage A, B and C, respectively. They are repeated multiple times and be serially connected. This configuration is denoted as IR 5-10-5 for reference. **Lower**: Schema of PolyNet. The design is based on the above schema with the modification that Inception residual units are replaced by PolyInceptions. The number of blocks in each stage might change for different models.

The corresponding architecture is shown in Fig. 4 (a). This unit comprises three paths, namely, an identity path, a first-order path that contains one Inception block, and a second-order path that contains two. The signals resulted from these paths will be summed together at the output end, before the *ReLU* layer. Here, the second-order path allows the input signal to go through deeper transformations before being merged back to the main path. This can considerably increase the expressive power of the unit.

It is noteworthy that along both the first-order and second-order paths, the outputs of the first Inception block are the same, that is, $F(\mathbf{x})$. Taking advantage of this observation, we can reduce the architecture in Fig. 4 (a) to a *cascaded form* as shown in Fig. 4 (b). These two forms are mathematically equivalent; however, the computational cost of the latter is only $2/3$ of the former. Note that the cascaded form can be reflected by an algebraic rewriting as $I + (I + F)F$. As such an extension embodies a *polynomial composition* of the *Inception* blocks, we call it a *PolyInception module*. The maximum number of Inception blocks along a path is referred to as the *order* of the module.

Figure 4 illustrates several PolyInception modules which differ in how the Inception blocks are composed and whether they share parameters. For convenience, we refer to them as *poly-2*, *mpoly-2*, and *2-way* respectively. The following discussion compares their characteristics.

- *poly-2*: $I + F + F^2$. This unit contains two Inception blocks, both denoted by $F$. This is a notational convention indicating that they share parameters. This design can increase the representation power without introducing additional parameters.

- *mpoly-2*: $I + F + GF$. The structure of this design is similar to *poly-2*, except that the two Inception blocks do *not* share parameters. Similar to *poly-2*, the computation of the first layer along both 1st- and 2nd-order paths can be shared (both denoted by $F$), and thus it can also be reduced into an equivalent cascaded form $(I+(I+G)F)$, as shown in Fig. 4 (c). Compare to *poly-2*, it possesses stronger expressive power. At the same time, however, the parameter size also doubles.

- *2-way*: $I + F + G$. This is a first-order PolyInception, with an extra first-order path incorporated into the unit. This construction is similar to the building blocks used in the *Multiple Residual Network* [1].

These exemplar designs can be further extended based on the relations between *operator polynomials* and network architectures. For instance, *poly-2* and *mpoly-2* can be extended to higher-order PolyInception modules, namely, *poly-3* $(I+F+F^2+F^3)$ and *mpoly-3* $(I+F+GF+HGF)$. Compared to their 2nd-order counterparts, the extended versions have greater expressive power, but also incur increased computational cost. Similarly, *2-way* can also be extended to *3-way*.

### 3.2. An Ablation Study

Before proceeding to the design of the entire network, we first study how the PolyInceptions can influence the overall performance. The study uses a trimmed down version of the Inception-ResNet-v2 *(IR-v2)* [25] as the baseline model. As shown in Fig. 5 (a), *IR-v2* comprises three stages A, B, C that operate on different spatial resolutions ($35\times35$, $17\times17$, and $8 \times 8$). These stages respectively contain 5, 10, and 5 Inception residual units. Hence, we denote the standard version of *IR-v2* as *IR 5-10-5*. This is a very large network. As we need to investigate a number of configurations in this study, to speed up the experiments, we scaled down the size a bit to *IR 3-6-3*, and used it as the baseline model.

In this study, we constructed a number of variant versions of *IR 3-6-3*. In each version, we choose one stage (from A, B, and C) and replace all the Inception residual units therein with one of the six PolyInception modules introduced above (*2-way*, *3-way*, *poly-2*, *poly-3*, *mpoly-2*, and *mpoly-3*). Hence, we have 18 configs in total (except the baseline). We trained networks for these configs, using the same training strategies and hyper-parameters, which will be detailed in Sec. 4.

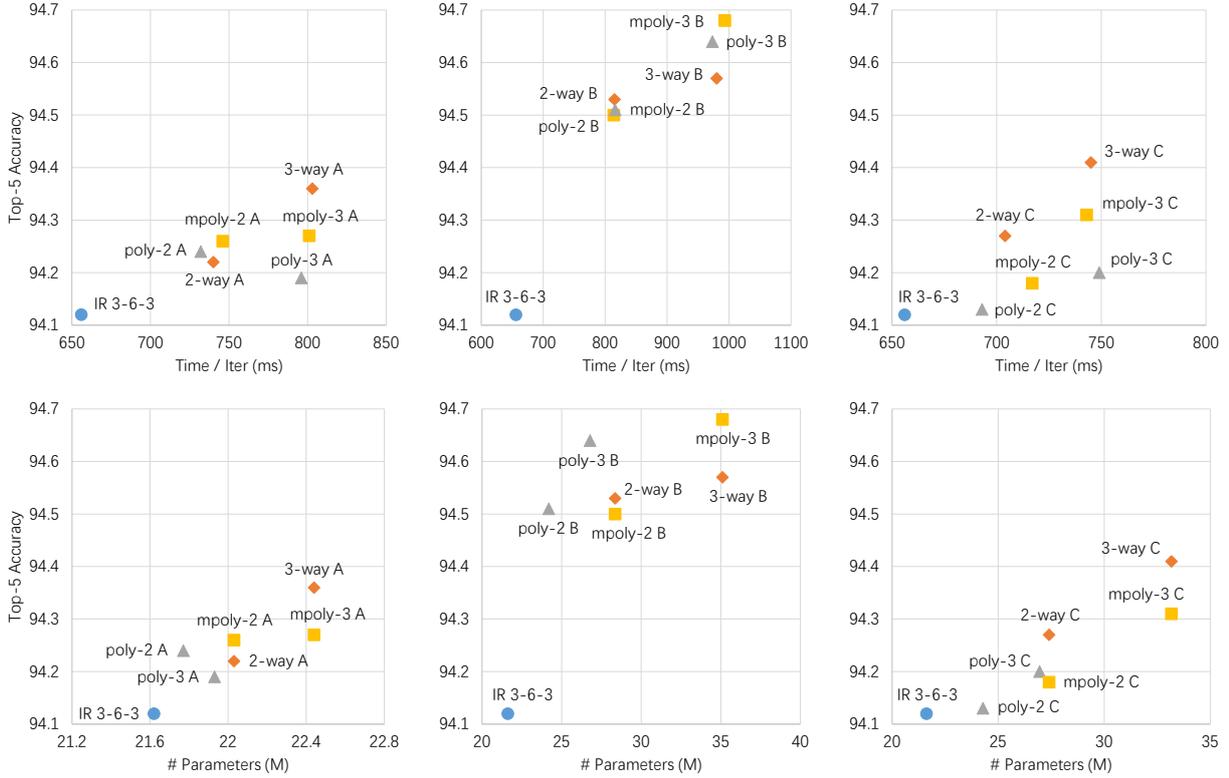

Figure 6: Comparison of different PolyInception configurations on 3-6-3 the scale. $y$-axis is the single crop top-5 accuracy on the ILSVRC 2012 validation set. The first row measures the accuracy v.s. computational cost, where the $x$-axis is the running time per iteration in real experiment. The second row shows the efficiency of parameters. The three columns are configurations that modify stage A, B and C, respectively. In these figures, the upper left area means higher accuracy with less time complexity or less parameter size.

Figure 6 compare the *cost efficiency* of different configs, by plotting them as points in *performance vs. cost* charts. Here, the performance is measured by the best top-5 single-crop accuracy obtained on the ILSVRC 2012 validation set; while the cost is measured in two different ways, namely, the run-time per iteration and the parameter size.

From the results, we can observe that replacing an Inception residual unit by a PolyInception module can always lead to performance gain, with increased computational cost. Whereas the general trend is that increasing the computational complexity or parameter size can result in better performance, the cost efficiencies of different module structures are significantly different. For example, enhancing stage B is the most effective for performance improvement. Particularly, for this stage, *mpoly-3* appears to be the best architectural choice for this purpose, while the runner-up *poly-3*. It is, however, worth noting that the parameter size of *poly-3* is only 1/3 relative to that of *mpoly-3*. This property can be very useful when the memory constraint is tight. For the other two stages, A and C, the performance gain is relatively smaller, and in these stages *k-way* perform slightly better than *mpoly* and *poly*. Such observations can provide useful guidance when designing a very deep network using PolyInception modules.

### 3.3. Mixing Different PolyInception Designs

In the study above, we replace all Inception blocks in a stage by PolyInception modules of certain configurations. While the study reveals interesting differences *between* different stages, a question remains: what if we use PolyInceptions of different configs *within* a stage?

We conducted an ablation study to investigate this problem. This study uses a deeper network, *IR 6-12-6*, as the baseline. This provides not only an opportunity to observe the behaviors in a deeper context, but also a greater latitude to explore different module compositions. Specifically, we focus on stage B, as we observe empirically that this stage has the greatest influence on the overall performance. Also, focusing on one of the stages allows us to explore more configurations given limited computational resources.

This study compares 5 configurations, namely, the *baseline IR 6-12-6*, three *uniform configs* that respectively replace the Inception blocks with *poly-3*, *mpoly-3*, and *3-way*, and a *mixed config*. Particularly, the *mixed config* divides the 12 units in stage B into 4 groups, replacing each group with a chain of PolyInception modules in three different configs, as shown in Fig. 7. We denote this structure by (*3-way* → *mpoly-3* → *poly-3*) ×4. Networks of all configs are trained under the same settings (detailed in Section 4).

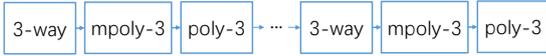

Figure 7: Mixed config interleaves *3-way*, *mpoly-3* and *poly-3*.

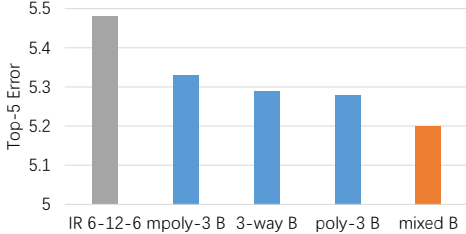

Figure 8: Performances of PolyNets on the 6-12-6 scale. $y$-axis is the single crop top-5 error on the ILSVRC 2012 validation set. The mixed structure uses (*3-way* → *mpoly-3* → *poly-3*) ×4 configuration. We can see that the mixed structure is slightly better than any of the three uniform configurations.

Figure 8 shows the comparative results. Again, we observed that replacing Inception residual units with PolyInception modules generally leads to notable performance gains, and that *poly-3* performs slightly better than the other two. A more noteworthy observation is that the *mixed config* outperforms all *uniform configs* by a considerable margin, with a comparable computational cost. This clearly shows that the benefit of enhanced structural diversity exists not only within each module but also across modules.

### 3.4. Designing Very Deep PolyNet

With all the studies presented above, we set out to design a very deep network, with an aim to take the state-of-the-art performance of a *single network* to a next level. This design follows the *basic* structure of the Inception ResNet-v2 [25], which comprises a *stem* at the bottom, and three stages on top that respectively operate on different spatial resolutions. The output of stage C will be connected with the class label via a fully connected layer and a softmax loss layer.

The objective of this design is to obtain optimal performance under a given computational budget – the network should be able to fit in a small cluster that contains 32 GPUs, each with $12GB$ RAM, when being trained with a batch size of $512$. Towards this objective, we carefully chose PolyInception modules or compositions thereof for each stage, with multiple factors taken into consideration, *e.g.* the balance between expressive power and cost, the structural diversity, and the training difficulty. Besides, the communication costs, which is determined by parameter size, is also important when training in a distributed environment.

At the end of the day, we obtain a final design, described as follows. Stage A comprises 10 *2-way* PolyInception residual units, Stage B comprises 10 mixtures of *poly-3* and *2-way* (20 modules in total), while Stage C comprises 5 mixtures of *poly-3* and *2-way* (10 modules in total). Note that while we take the studies presented above as references, some modifications are made to fit the network on the GPUs, reduce the communication cost, and maintain a sufficient depth. For example, we change the *3-way* and *mpoly-3* modules used in the best-performing mixed structure in the study above to *2-way* and *poly-3*.

## 4. Training PolyNets

We trained our networks on the training set of the ILSVRC image classification benchmark [19], and evaluated their performance on its validation set. The task is to classify each given image to one of the 1000 categories. Here, the training set contains about $1.28M$ images, while the validation set contains $50K$. Below, we describe the technical settings in our training procedures.

**Data augmentation.** To improve the generalization performance, we adopt the multi-area augmentation scheme presented in [26]. Specifically, for each training image, we randomly generate a crop with the constraints that the crop area ranges from $8\%$ and $100\%$ of the whole area and that the aspect ratio is between $3/4$ and $4/3$. The crop is then resized to a standard size $299 \times 299$ with bilinear interpolation, and flipped horizontally with a chance of $50\%$. We provide $331 \times 331$ input to the Very Deep PolyNet, which is slightly larger.

**Training settings.** We use RMSProp [29] in our training, with decay set to $0.9$ and $\epsilon = 1.0$. Batch normalization is used following the settings in [25]. The learning rate is initialized to $0.45$, and will be scaled by a factor of $0.1$ for every $160K$ iterations. All training procedures were terminated after $560K$ iterations – throughout the entire process, the learning rate will be decreased by three times.

We train our models on servers with 8 *Nvidia Titan X* GPUs, with all computation performed with 32-bit single precision. Models of different scales are trained on different numbers of nodes. All models based on *IR 3-6-3*, *e.g.*, those studied in Sec. 3.2, are trained on single nodes, using mini-batches of size $256$. Those based on *IR 6-12-6*, *e.g.*, those studied in Sec. 3.3, are trained on double-node clusters, using mini-batches of size $256$. Very large models, including *ResNet-500*, *Very Deep Inception-ResNet*, and the *Very Deep PolyNet* as described in Sec. 3.4, are trained on four-node clusters, using mini-batches of size $512$. In all settings, each mini-batch is evenly dispatched to all GPUs. Since batch normalization is performed locally within each GPU, the convergence rate and also the final performance would be adversely affected if the number of samples assigned to each GPU is too small. That's why we use a larger batch size when training on four-node clusters – this ensures that each GPU processes at least 16 samples at each iteration.

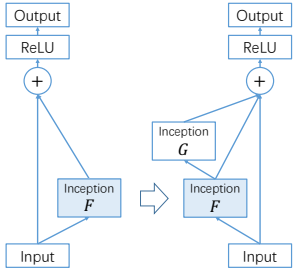
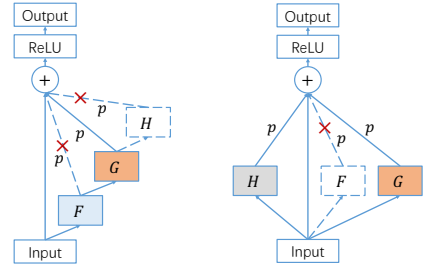

Figure 9: Two examples to illustrate initialization-by-insertion. **Left:** When replacing a residual unit by a PolyInception module, we retain the parameters of the 1st-order block $F$, and randomly initialize $G$, the newly inserted block. **Right:** When doubling the depth of a network, we insert new units in an *interleaved* manner, that is, inserting new units *between* original units. We found this more effective than stacking all new units together.

Figure 10: Examples of stochastic paths. A PolyInception module will be converted into a smaller but distinct one when some paths are dropped. **Left:** from $I + F + GF + HGF$ to $I + GF$. **Right:** from $I + F + G + H$ to $I + G + H$.

**Initialization.** For very deep networks, random initialization may sometimes make the convergence slow and unstable. Initializing parts of the parameters from a smaller pre-trained model has been shown to be an effective means to address this issue in previous work [9, 21]. Inspired by these efforts, we devise a strategy called *initialization by insertion* to initialize very deep networks, as illustrated in Fig. 9. We found that this strategy can often stabilize and accelerate the training, especially in early iterations.

**Residual scaling.** As observed in [25], simply adding the residuals with the input often leads to unstable training processes. We observed the same and thus followed the practice in [25], namely, scaling the residual paths with a *dampening factor* $\beta$ before adding them to the module outputs. Take *2-way* for instance. This modification is equivalent to changing the operator from $I + F + G$ to $I + \beta F + \beta G$. Particularly, we set $\beta = 0.3$ in our experiments.

**Stochastic paths.** Very deep networks tend to overfit in later stage of training, even on data sets as large as ImageNet. Strategies that randomly drop parts of the network, such as Dropout [23], DropConnect [31], Drop-path [17], and Stochastic Depth [13], have been shown to be effective in countering the tendency of overfitting.

Inspired by such works, we introduce a technique called *stochastic paths*, which randomly drop some of the paths in each module with pre-defined probabilities, as illustrated in Fig. 10. It is worth noting that the diverse ensemble implicitly embedded in the PolyNet design is manifested here – with random path dropping, a PolyInception module may be turned into a smaller but distinct one. At each iteration, a different sub-network that combines a highly diverse set of modules will be made visible to the training algorithm and the associated parameters updated accordingly. This way can reduce the co-adaptation effect among components, and thus result in better generalization performance.

Note that we found that applying stochastic paths too early can hinder the convergence. In our training, we do this adaptively – activating this strategy only when serious overfitting is observed, where the dropping probabilities from bottom to top are increased linearly from 0 to 0.25.

## 5. Overall Comparison

We compared our Very Deep PolyNet (Sec. 3.4) with various existing state-of-the-art architectures. It is worth noting that the computation cost of the Very Deep PolyNet is significantly larger than Inception-ResNet-v2 or the deepest version of publicly available ResNet, namely ResNet-200 by Facebook. To get a persuasive comparison, we built a Very Deep Inception-ResNet by stacking more residual units in Inception-ResNet-v2[3] with 20, 56, 20 units respectively in stages A, B, and C. We also extended the ResNet to ResNet-269 and ResNet-500. The Very Deep Inception-ResNet and the ResNet-500 both have similar computation costs to our Very Deep PolyNet. We evaluate both single crop and multi-crop classification errors of these models on the whole ILSVRC 2012 validation set.

The multi-crop evaluation we carried out was based on *top-k pooling* [32]. Specifically, we generated crops as described in [26], using 8 scales and 36 crops for each scale. Then we did *top-30% pooling* in each scale, *i.e.* averaging the top 30% scores among crops for each class independently, and finally averaged scores across all 8 scales.

We first examine Fig. 11. In this experiment, we gradually scaled the original Inception-ResNet-v2 from *IR 5-*

---

[3]We did not use auxiliary loss or label-smoothing regularization as in [25, 27]. This leads to a small gap (0.2% for top-5 accuracy) between our implementation and that reported in [25]. To ensure fair comparisons, all results reported in this paper, except some items Table 1, are based on our own implementations.

Table 1: Single model performance with single crop and multi-crop evaluation on ILSVRC 2012 validation set and test set.
† indicates that the model was trained by us. Error rate followed by ‡ means that it was evaluated on the non-blacklisted subset of validation set [25], which may lead to slightly optimistic results.

| Network | Single-crop on validation set | | Single-crop on test set | Multi-crop on validation set | |
| --- | --- | --- | --- | --- | --- |
| | Top 1 Error (%) | Top 5 Error (%) | Top 5 Error (%) | Top 1 Error (%) | Top 5 Error (%) |
| ResNet-152 [9] | 22.16 | 6.16 | - | 19.38 | 4.49 |
| ResNet-152† | 20.93 | 5.54 | 5.50 | 18.50 | 3.97 |
| ResNet-269† | 19.78 | 4.89 | 4.82 | 17.54 | 3.55 |
| ResNet-500† | 19.66 | 4.78 | 4.70 | 17.59 | 3.63 |
| Inception-v4 [25] | 20.0 ‡ | 5.0 ‡ | - | 17.7 | 3.8 |
| Inception-ResNet-v2 [25] | 19.9 ‡ | 4.9 ‡ | - | 17.8 | 3.7 |
| Inception-ResNet-v2† | 20.05 | 5.05 | 5.11 | 18.41 | 3.98 |
| Very Deep Inception-ResNet† | 19.10 | 4.48 | 4.46 | 17.39 | 3.56 |
| Very Deep PolyNet† | **18.71** | **4.25** | **4.33** | **17.36** | **3.45** |

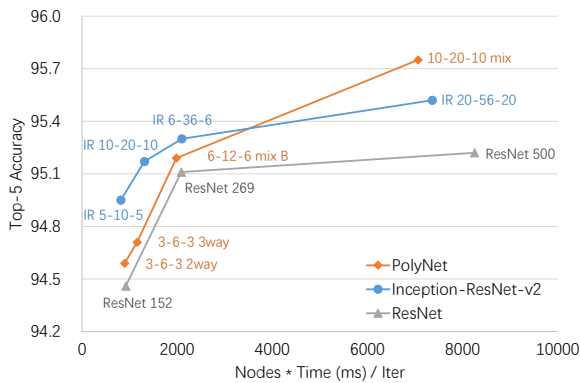

Figure 11: With PolyInception, large performance gain can still be observed on very deep network. The results suggest that structural diversity scales better than a mere deeper network.

*10-5* to the Very Deep Inception-ResNet denoted as *IR 20-56-20*. The top-5 accuracies of these models are plotted. We similarly plotted a curve that depicts the performance of our model of varying complexity, from PolyNet 3-6-3 with *2-way* structure to the final Very Deep PolyNet with *mixed config*. Note that we applied adaptive stochastic depth (Sec. 4) to ameliorate overfitting for very deep models. Fig. 11 shows several important observations: (1) Deeper structure is more desirable when the computational budget is limited (below 3000 ms/iter in Figure 11). (2) As the budget increases, the gain of structural diversity grows more rapidly. Particularly, beyond 4000 ms/iter, structural diversity could bring larger performance gain to very deep network given the same computational budget, as compared to simply stacking more units. (3) The architectures with Inception blocks generally perform better than the standard ResNet architectures (stacks of $3 \times 3$ convolutional layers) under the same computational budget. This also partly corroborates the significance of structural diversity. Overall, these observations reveal the complementary roles of the depth and structural diversity, which could facilitate future deep architectural design.

Table 1 summarizes the top-1 and top-5 errors of single/multi-crop evaluations. It shows that the proposed Very Deep PolyNet not only outperforms state-of-the-art ResNet and Inception-ResNet-v2, but also their very deep variants, in both the validation and test sets of ImageNet. In particular, we obtain top-5 error rates $4.25\%$ and $3.45\%$, respectively on the single/multi-crop settings on the validation set, and $4.33\%$ on the single-crop setting on the test set. This improvement is notable compared to the best published single-network performance obtained by IR-v2, *i.e.* $4.9\%$ and $3.7\%$ on single-crop and multi-crop settings. It is also worth noting that the multi-crop performance of ResNet-500 is slightly lower than ResNet-269. This shows that deepening the network without enhancing structural diversity may not necessarily lead to positive gains.

## 6. Conclusion

In this paper, we explore a novel direction, *structural diversity*, to design deep networks. On this direction, we proposed a family of modules named *PolyInception* to extend residual style networks. We conducted systematic experiments to explore the behavior of *PolyInception*. Based on these studies, we designed a Very Deep *PolyNet*. The model outperforms Inception-ResNet-v2, the state of the art, and even its deeper variants. Our study shows that enhancing *structural diversity* along with going deep can lead to further performance improvements.

**Acknowledgements** This work is partially supported by the SenseTime Group Limited, the Hong Kong Innovation and Technology Support Programme, and Early Career Scheme (ECS) grant. The authors would also like to thank all the members of the *CU-DeepLink* [4] Team for the hard work in the ImageNet 2016 competition. Special thanks goes to Tong Xiao for providing trained ResNets.